\documentclass[]{article}

\usepackage{lscape,caption}
\usepackage{amssymb}
\usepackage{amsmath}
\usepackage{mathtools, cuted}
\usepackage{graphicx, subfigure}
\usepackage{url}
\usepackage{algorithm}
\usepackage{algpseudocode}
\usepackage{color}
\usepackage{booktabs}
\usepackage{stmaryrd}
\usepackage{epstopdf}
\usepackage{pdflscape}
\usepackage{stmaryrd}
\usepackage{color}
\usepackage{listings}

\definecolor{codegreen}{rgb}{0,0.6,0}
\definecolor{codegray}{rgb}{0.5,0.5,0.5}
\definecolor{codepurple}{rgb}{0.58,0,0.82}
\definecolor{backcolour}{rgb}{0.95,0.95,0.92}

\lstdefinestyle{mystyle}{
	backgroundcolor=\color{backcolour},   
	commentstyle=\color{codegreen},
	keywordstyle=\color{magenta},
	numberstyle=\tiny\color{codegray},
	stringstyle=\color{codepurple},
	basicstyle=\ttfamily\footnotesize,
	breakatwhitespace=false,         
	breaklines=true,                 
	captionpos=b,                    
	keepspaces=true,                 
	numbers=left,                    
	numbersep=5pt,                  
	showspaces=false,                
	showstringspaces=false,
	showtabs=false,                  
	tabsize=2
}

\title{mldr.resampling: Efficient Reference Implementations of Multilabel Resampling Algorithms}

\author{Antonio J. Rivera \and
	Miguel A. D\'avila \and
	David Elizondo \and
	Mar\'ia J. del Jesus \and
	Francisco Charte\thanks{Corresponding author: Francisco Charte, Computer Science Department, A3-241 Universidad de Ja\'en, Campus Las Lagunillas, 23071 Ja\'en, Spain.}}

\begin{document}

\maketitle

\begin{abstract}
Resampling algorithms are a useful approach to deal with imbalanced learning in multilabel scenarios. These methods have to deal with singularities in the multilabel data, such as the occurrence of frequent and infrequent labels in the same instance. Implementations of these methods are sometimes limited to the pseudocode provided by their authors in a paper. This Original Software Publication presents \texttt{mldr.resampling}, a software package that provides reference implementations for eleven multilabel resampling methods, with an emphasis on efficiency since these algorithms are usually time-consuming.
\end{abstract}

%\end{frontmatter}

\section{Introduction}

MultiLabel Learning (MLL)~\cite{Charte:SB-MLC} is one of the most common machine learning tasks today. It is based on the idea that each data sample is associated with a certain subset of labels. The full set of labels can be large, in many cases even having more labels than input features. As a result, it is common for some labels to occur in only a few samples, while others occur much more frequently. The label imbalance~\cite{Charte:Neucom13} in MLL is almost always present, and it is a serious obstacle to training good classifiers.

Class imbalance is a very well-known problem in traditional learning tasks such as binary and multiclass classification. Hundreds of articles~\cite{Luque2019TheIO,sun2009classification,haixiang2017learning}, conference papers~\cite{Menon2013OnTS} and books~\cite{HeBook:2013} have been devoted to studying it and proposing possible solutions. The most popular are data resampling, cost-sensitive learning and mixtures of these approaches~~\cite{kotsiantis2006handling,mohammed2020machine}. However, imbalanced learning in the MLL field presents some specific aspects that make it more difficult to deal with this problem.

Resampling algorithms, which can generate new samples associated with underrepresented labels or remove those that are overrepresented, are a model-independent solution to imbalance. Several such algorithms adapted to multilabel datasets (MLD) have been proposed in recent years~\cite{Charte:Neucom13,sadhukhan2019reverse,Charte:MLSMOTE,liu2022multi,pereira2020mltl,Charte:IDEAL14,Charte:REMEDIAL}. However, reference implementations are not yet available for all of them.

Several of these proposals rely heavily on finding nearest neighbors between each data sample and all the others in the dataset. This is a computationally expensive task. It will be desirable to take advantage of modern hardware and speed up this work through parallelization. This will make the time needed to resample some MLDs more affordable for everyone.

The \texttt{mldr.resampling} software package is a new member of the \textit{mldr ecosystem}, which also includes the \texttt{mldr}~\cite{Charte:mldr} and the \texttt{mldr.datasets}~\cite{charte2018tips} packages. All of them are written in R language and are available from CRAN (\textit{Comprehensible R Archive Network}), the official repository of R software. The new package provides implementations of a dozen of resampling algorithms for MLDs. 

This document is structured as follows. Section~\ref{Sec.Background} sets the context from which the package originates. The software itself, its structure, functionalities, and implementation details are described in Section~\ref{Sec.Software}. Several illustrative usage examples are then provided in Section~\ref{Sec.Examples}. Finally, some conclusions are drawn in Section~\ref{Sec.Conclusions}.

\section{Background}\label{Sec.Background}

Labeled data is used everyday to create models through supervised algorithms. These usually assume that each data pattern has only one label. However, this is not always the case. Beyond the scenarios of binary learning ---e.g. an email is considered as spam or it is not--- and multiclass learning ---e.g. a flower is cataloged into one of a set of species--- there are others such as MLL~\cite{Charte:SB-MLC}.  This is considered a non-standard learning situation~\cite{Charte:Nonstandard}, among others such as multiview learning~\cite{multi-view} and multiinstance learning~\cite{zhou2004multi}.

Let $X^1,\dots X^f$ be the domains of the $f$ features in a dataset and $L$ be the set of distinct labels. The \textit{i-th} data pattern $I_i$ in an MLD is defined as~(\ref{Eq.MLL}).

\begin{equation}
	I_i = (X_i, Y_i) \mid X_i \in X^1\times X^2\times \dots\times X^f, Y_i \subseteq L .
	\label{Eq.MLL}
\end{equation}

In MLL $Y_i$ can be any subset of $L$, instead of just one label as in binary and multiclass learning. This also includes the empty set and the full set of labels. Given an MLD $D$ with instances $x_i$, the goal of MLL is to find a model such as the one in~(\ref{Eq.MLLf}) that is able to predict $Y'_i$, i.e. the subset of labels applicable to each instance.

\begin{equation} 
	Y'_i=f(x_i), Y'_i \in L  . 
	\label{Eq.MLLf} 
\end{equation}

MLL has been applied is disparate fields, including the processing of narratives related to aviation safety~\cite{robinson2018multi}, content-based retrieval of remote sensing images~\cite{dai2018novel}, predicting the toxicity effects of chemicals~\cite{liu2018integrated}, and the automatic tagging of posts in forums~\cite{QUINTA}, among many others. Several reviews of all these works and other MLL techniques are available~\cite{Zhang:2013,ReviewVentura,TutorialVentura}.

The data used for training a classifier is often imbalanced. As a result, the class labels that are assigned to each instance are not equally represented. For binary datasets~\cite{Japkowicz:2002} and to a lesser extent for multiclass ones~\cite{Alberto:2013}, this is a well studied problem. In order to know the level of imbalance of the datasets, a measure called \textit{imbalance ratio}) (\textit{IR})~\cite{Japkowicz:2002} is used. 

\begin{equation}
	\textit{IR} = \dfrac{\# samples~majority~class}{\# samples~minority~class}
	\label{Eq.IR}
\end{equation}

In a binary scenario, this ratio is computed as shown in (\ref{Eq.IR}). The greater the proportion of instances associated with the most common class as opposed to the minority one, the higher the \textit{IR} will be. When there are more than two classes or labels, individual \textit{IRs} are obtained pairwise, then an average \textit{IR} ---usually noted as \textit{MeanIR}--- is calculated.

Traditionally, techniques such as data resampling, cost-sensitive learning, and algorithm-specific adaptations have been used to deal with imbalanced classification~\cite{Lopez:2013}. The former have the advantage of being model independent, i.e. they do not require changes to the model to improve the learning process. 

\subsection{Highly imbalanced labels}

\begin{figure}[h!]
	\includegraphics[width=\linewidth]{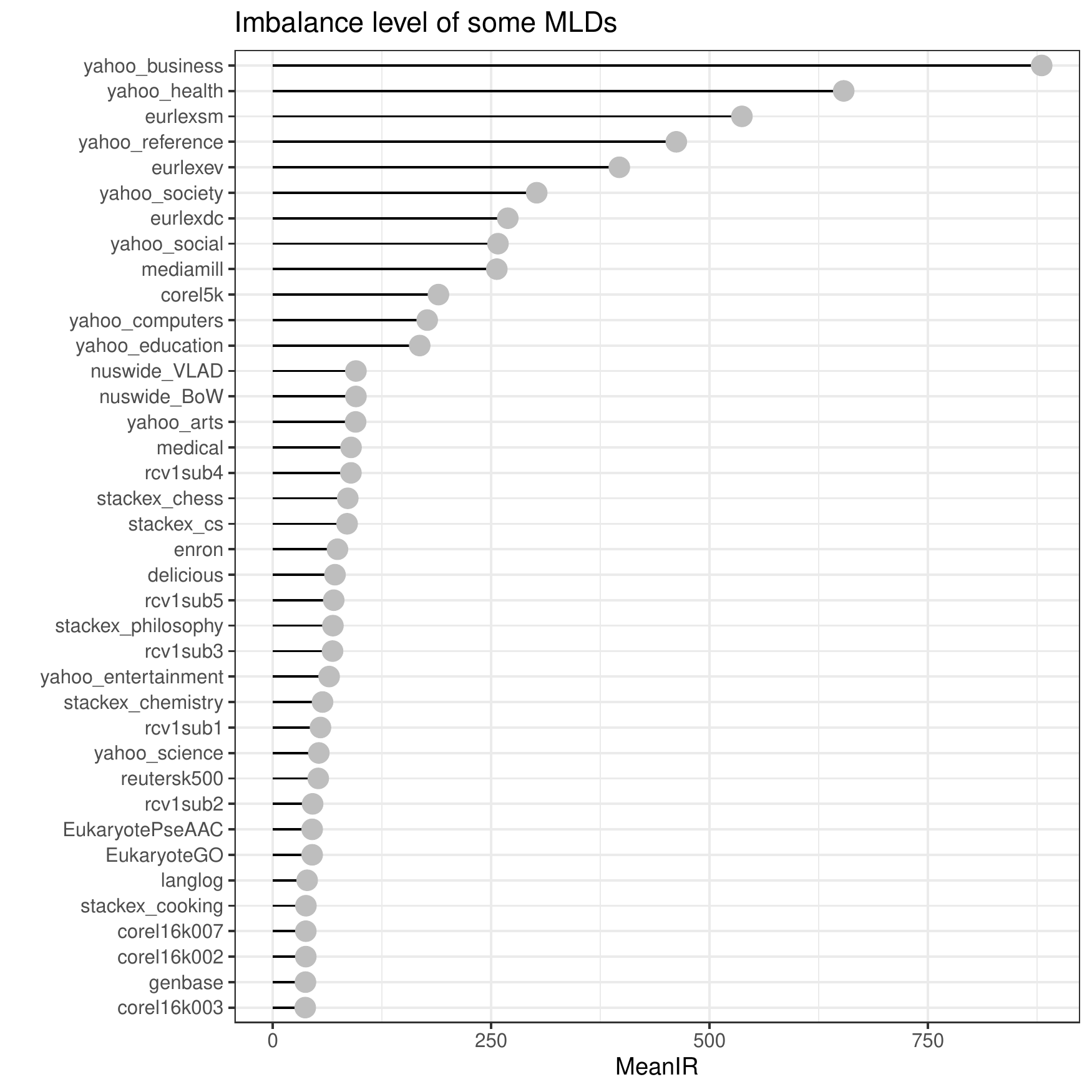}
	\caption{Mean imbalance ratio of MLDs obtained from the Cometa dataset repository. Datasets having MeanIR below 35 or above 1000 have been excluded.}
	\label{Fig.ImbalanceLevels}
\end{figure}

Imbalance in the MLL domain presents some additional difficulties. While in the binary and multiclass scenarios having $\textit{MeanIR}>10$ is generally considered a high level of imbalance, in MLL it is quite common. In fact, only a few popular MLDs\footnote{\textit{IR} of most publicly available MLDs, among other common metrics, are available in the Cometa multilabel dataset repository: \url{https://cometa.ujaen.es/datasets}.} have a $\textit{MeanIR}\le 10$, as can be seen in Figure~\ref{Fig.ImbalanceLevels}. Many of them have $\textit{MeanIR}\in [30, 100]$, a few suffer from $\textit{MeanIR}\in [150, 500]$ and there are some extreme cases with $\textit{MeanIR}>1000$.

This means that there are rare labels that occur once in several hundreds of the common ones. It is therefore extremely difficult to learn a model that correctly handles these low frequency labels. The problem can be exacerbated when some MLL techniques, such as binary relevance~\cite{Godbole}, are applied, since an independent model is learned for each label against all the others.

\subsection{Coupling of frequent and rare labels}

Most of the popular solutions for dealing with data imbalance when training models, derived from the standard scenarios (binary and multiclass classification), cannot be directly applied in the MLL case due to the unique nature of the data samples. In the first case, it is possible to collect all the samples associated with the minority class and generate some synthetic instances, so that the imbalance ratio is reduced. Similarly, some of the instances corresponding to the majority class could be removed. Since each data pattern corresponds to only one label, it is quite easy to apply data resampling methods.

\begin{figure}[h!]
	\includegraphics[width=\linewidth]{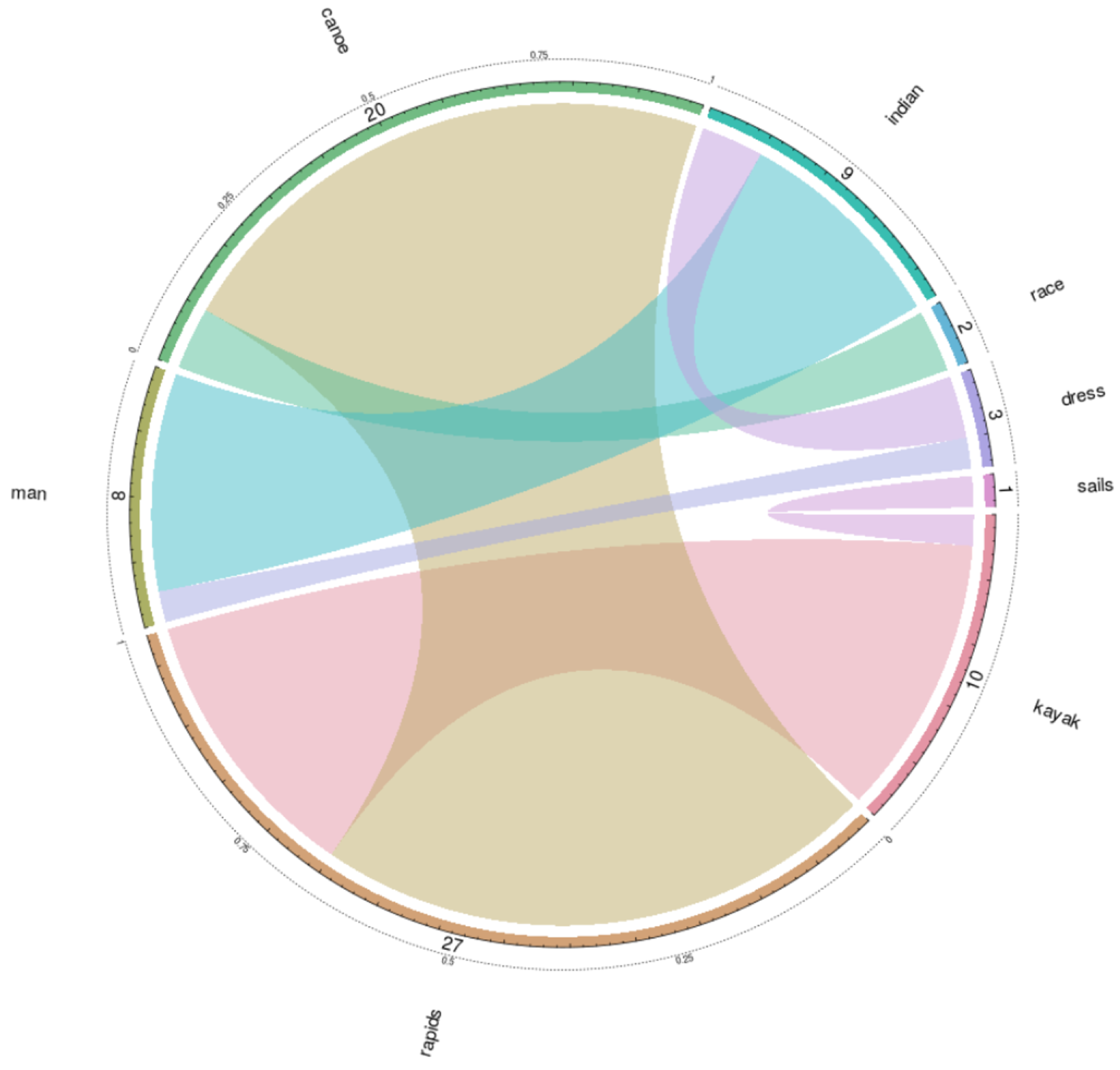}
	\caption{This concurrence diagram shows how all the instances of three minority labels, in the right side of the plot, always appear in instances having one or more majority labels.}
	\label{Fig.Coupling}
\end{figure}

In MLDs, each instance has multiple labels associated with it, so when a sample is removed or produced, all of its labels are affected, not just the most common or rarest label. In addition, it is quite normal in MLDs for the less frequent labels to appear coupled to the majority ones~\cite{Charte:NeucomDifficultLabels} in the same instances (see Figure~\ref{Fig.Coupling}). This casuistic is measured by means of the \textit{SCUMBLE} metric introduced in~\cite{Charte:NeucomDifficultLabels}. As a consequence, an increase in the number of samples containing rare labels also implies an increase of the most common ones, and the \textit{IR} remains unchanged.

\subsection{Computational cost}

Although high imbalance levels and the coupling of rare and frequent labels are the main obstacles to learning with imbalanced MLDs, there is another fact to consider: the huge computational cost imposed by some of the proposed solutions.

First of all, MLDs sometimes have hundreds if not thousands of labels. This means that there is no single majority or minority label, but sets of them. Tuning the frequency of all instances associated with these sets is more costly than dealing with just one minority class as in binary or multiclass learning.

In addition, there are proposed algorithms for resampling imbalanced MLDs that rely heavily on nearest neighbor search. Computing the distances between all the samples in a dataset is already computationally expensive. In the case of MLDs, the cost is higher because of the usually large number of features ---many MLDs come from text document classification--- and also the need to repeat the search for each minority or majority label to be resampled.

\subsection{Resampling algorithms}

The methods proposed in the literature to deal with imbalance in the MLL field are mostly based on data resampling. Many of the existing MLL methods are built as ensembles of simpler models. This is the case of~\cite{Tsoumakas4,HOMER,Read,read2008multi,read2015scalable} among many others. Introducing cost-sensitive learning or other algorithmic modifications to improve learning from imbalanced data is not an easy task. On the contrary, a data resampling algorithm can be applied in a preprocessing step, regardless of the MLL model used in the end.

\begin{table}[h!]\centering
	\small\def\arraystretch{1.5}
	\begin{tabular}{lp{5.2cm}c}
		\toprule
		\textbf{Name} & \textbf{Description} & \textbf{Ref.} \\
		\midrule
		LPROS &  Random oversampling of labelsets & \cite{Charte:Neucom13}\\
		MLROS & Random oversampling of one minority label & \cite{Charte:Neucom13}\\
		MLRkNNOS & Reverse-nearest neighborhood based oversampling & \cite{sadhukhan2019reverse}\\
		MLSMOTE & Synthetic oversampling based on neighborhood similarity & \cite{Charte:MLSMOTE}\\
		MLSOL & Oversampling based on local label imbalance & \cite{liu2020synthetic}\\
		LPRUS & Random undersampling of labelsets & \cite{Charte:Neucom13}\\
		MLRUS & Random undersampling of one majority label & \cite{Charte:Neucom13}\\
		MLTL  &  Tomek Link undersampling & \cite{pereira2020mltl}\\
		MLUL & Undersampling based on local label imbalance & \cite{liu2020synthetic}\\
		MLeNN & Heuristic undersampling of majority labels & \cite{Charte:IDEAL14}\\
		REMEDIAL & Decoupling of minority and majority labels & \cite{Charte:NeucomDifficultLabels}\\
		\bottomrule
	\end{tabular}
	\caption{Some of the most popular resampling algorithms for MLDs and reference to the papers where they were introduced.}
	\label{Tbl.Methods}
\end{table}

Table~\ref{Tbl.Methods} lists most of the MLL resampling methods proposed in the literature. There are five oversampling algorithms, five more that perform undersampling, and one that takes a different approach by decoupling instances where rare and frequent labels coexist. The rightmost column gives the reference to the paper where each algorithm was introduced. We have used the pseudo-code in these papers as a guide to implementing each of them. Most of them have been reviewed in a recent paper~\cite{TAREKEGN2021107965}.

\section{Software framework}\label{Sec.Software}

This section describes the functionality provided by the software, as well as some internals about how it works. Its dependencies, i.e. other software packages on which it depends, are also listed.

\subsection{Software architecture}

Our \texttt{mldr.resampling} package works with multilabel datasets. The infrastructure for this is supplied by the \texttt{mldr} package~\cite{Charte:mldr} as an R S3 class, also called \texttt{mldr}. S3 is an object-oriented mechanism in the R language that relies on generic functions. So most of the functions exported by \texttt{mldr.resampling} take a \texttt{mldr} object as a parameter. Users would load their MLDs using the facilities offered by \texttt{mldr} and get a \texttt{mldr} object as a result. These objects are later used as inputs to the methods in \texttt{mldr.resampling}.

Each one of the resampling algorithms shown in the Table~\ref{Tbl.Methods} is implemented in \texttt{mldr.resampling} as an independent method. This allows the users to experiment with the algorithm(s) they are interested in without any additional software layers.

As well as applying an algorithm to a MLD, the package also considers the possibility of running several methods to compare them. To achieve this goal, a function is provided that automates the procedure of calling each of the algorithms with the same MLD and the correct parameters.

\subsection{Software functionalities}

The main contributions of this new software package are as follows:

\begin{itemize}
	\item Reference implementations of eleven resampling algorithms for MLD, with source available after installing the package.
	\item A unified interface that facilitates the application of several of these algorithms to an MLD, automating several of the steps required to compare them.
	\item Built-in optimizations to speed up the nearest neighbor search through task parallelization and neighbor caching.
\end{itemize}

These functionalities can be accessed in different ways depending on the user's needs. To apply any of the algorithms to one MLD the following methods are available:

\begin{itemize}
	\item \texttt{LPROS(mld, pc)}: Resamples the \texttt{mld} given as the first input until a \texttt{pc}\% increase in the number of samples is reached.
	\item \texttt{LPRUS(mld, pc)}: Resamples the \texttt{mld} given as the first input until a \texttt{pc}\% decrease in the number of samples is reached.
	\item \texttt{MLeNN(mld, threshold, k)}: Removes samples in the \texttt{mld} with majority labels whose differences with their \texttt{k} nearest neighbors are above the \texttt{threshold}, following the ENN rule~\cite{wilson1972asymptotic}.
	\item \texttt{MLRkNNOS(mld, k)}: Resamples the \texttt{mld} producing new instances based on the \texttt{k} reverse-nearest neighbors.
	\item \texttt{MLROS(mld, pc)}: Resamples the \texttt{mld} by cloning random instances in which minority labels appear up to a point at which a \texttt{pc}\% increase is reached.
	\item \texttt{MLRUS(mld, pc)}: Resamples the \texttt{mld} given as first input removing random instances in which majority labels appear up to a point at wich a \texttt{pc}\% decrease is reached.
	\item \texttt{MLSMOTE(mld, k)}: Generates synthetic instances in which one or more minority labels appear, producing characteristics from \texttt{k} nearest neighbors.
	\item \texttt{MLSOL(mld, pc, k)}: Resamples the \texttt{mld} creating new instances on difficult regions of the instance space, using local information of \texttt{k} neighbors.
	\item \texttt{MLTL(mld, threshold)}: Removes instances in the \texttt{mld} based on Tomek Links~\cite{tomek1976two} strategy.
	\item \texttt{MLUL(mld, pc, k)}: Resamples the \texttt{mld} removing instances on difficult regions of the instance space, using local information of \texttt{k} neighbors.
	\item \texttt{REMEDIAL(mld)}: Decouples highly imbalanced labels occurring in the same samples by splitting each into two instances, one with the minority labels and another with the majority labels.
\end{itemize}

It is quite common to be interested in trying different resampling approaches on the same dataset. This allows several algorithms to be compared and the best one chosen for a particular MLD. Although the aforementioned methods follow different strategies to create or remove data samples, many of them rely on retrieving information from their neighbors. Since most MLDs have hundreds or even thousands of features, as well as a similar number of instances, finding \texttt{k} nearest neighbors for every instance has a huge impact on the time needed to run them.

The \texttt{mldr.resampling} package provides an additional method ---named \texttt{resample(mld, algorithms)}--- capable of executing all methods specified by the \texttt{algorithms} parameter on the same \texttt{mld}. This not only automates work that would otherwise require multiple function calls, but also improves the efficiency of running these methods through a caching mechanism. Once the nearest neighbor information is obtained, it is cached and passed as input to the individual methods, so that this task does not consume any time.

Another feature built into the package, accessible to the user via the \texttt{setParallel()} function, is the ability to distribute computations across all the cores in the system, instead of using just one as usual. Limiting the number of cores used by algorithms is also possible by calling the \texttt{setNumCores()} function.

The current set of algorithms included in the package is expected to grow in the coming period as new proposals are made available with the minimum guidelines needed to implement them. The structure of the code makes makes this process easy.

\subsection{Implementation details}

All methods have been implemented in pure R, so that the software package does not need to be compiled, and the source code is available to any R practitioner interested in digging into the code. Each method has been carefully written according to the instructions given in the respective paper. Where the original source code was available, the results of the new implementation were compared with those of the original to ensure consistency.

Although having all the methods available in a language like R makes them accessible to a large group of researchers, R is certainly not the most efficient language when it comes to execution speed. For this reason, the \texttt{mldr.resampling} package has been written with efficiency in mind through the following approaches:

\begin{itemize}
	\item The garbage collection process of most dynamic languages can have a large impact on execution times. Since most objects in R are immutable, growing certain data structures ---such as vectors and matrices--- means allocating new blocks of memory, copying old data, and freeing previously allocated memory. The \texttt{mldr.resampling} package takes advantage of the language's vectorization facilities whenever possible, avoiding loops to minimize the time spent on garbage collection.
	
	\item Harnessing the power of today's multi-core processors is essential to reduce execution time. However, few languages provide a standard way of doing this so that it can be used on all operating systems. The \texttt{mldr.resampling} code is designed so that some tasks can be run sequentially or in parallel, depending on whether the \texttt{parallel} R package is installed. By default, parallelization is not active, so the \texttt{parallel} package is not required. If it is available\footnote{In latest versions of R this package is included in the base installation.}, the user can enable parallelism simply by calling the \texttt{setParallel()} function with the \texttt{TRUE} value as a parameter.
	
	\item Many of the resampling methods in the package are based on nearest neighbor search. In their original implementation, this is the task that consumes most of the running time. For numerical attributes, finding the $k$ nearest neighbors for a given data sample means computing distances, almost always Euclidean, with all other instances in the MLD. If the dataset contains nominal attributes, the distances are even more expensive to obtain by VDM (\textit{Value Difference Metric})~\cite{stanfill1986toward}. All methods available in the \texttt{mldr.resampling} package optionally take two additional parameters, \texttt{neighbors} and \texttt{tableVDM}, which allow reusing these calculations. This caching method is automatically used when the \texttt{resample()} function is called to run more than one algorithm on the same data.
\end{itemize}
 
\section{Illustrative examples}\label{Sec.Examples}

The latest stable\footnote{The development version of the package is hosted on GitHub: \url{https://github.com/madr0008/mldr.resampling}.} version of the \texttt{mldr.resampling} package is available on CRAN, so it can be installed like any other R package by calling \texttt{install.packages()}, as shown below.  Once installed, the \texttt{library()} function will load the package and display a message to the user about how to enable parallel processing.

\begin{lstlisting}[language=R]
> install.packages("mldr.resampling")
> library(mldr.resampling)
Enter setParallel(TRUE) to enable parallel computing
\end{lstlisting}

Some MLDs are needed to run the resampling algorithms. As \texttt{mldr.resampling} has been installed, this implies that the \texttt{mldr} package has also been installed. The command \texttt{data(package="mldr")} will list the MLDs available in this package. Any other MLD can be fetched from the file system or downloaded from the Cometa repository using the functions provided by the \texttt{mldr.datasets} package.

By means of the same \texttt{data()} function the \texttt{emotions} MLD is loaded into memory and some measurements, such as the number of instances, average \textit{IR} and \textit{SCUMBLE}, are obtained:

\begin{lstlisting}[language=R]
> data(emotions, package="mldr")
> emotions$num.instances
[1] 593

> emotions$meanIR
[1] 1.478068

> emotions$scumble
[1] 0.01095238
\end{lstlisting}

This dataset is slightly imbalanced, with $\textit{MeanIR}=1.4781$, and the coupling of minority and majority labels is also bearable, with $\textit{SCUMBLE}=0.0109$. Nevertheless, it may benefit from the application of some resampling algorithm. A first option could be the simple LPROS oversampling method:

\begin{lstlisting}[language=R]
> lpemotions <- LPROS(emotions, P=25)
> lpemotions$measures[c("num.instances", "meanIR", "scumble")]
$num.instances
[1] 741

$meanIR
[1] 1.355174

$scumble
[1] 0.007740464
\end{lstlisting}

The call to the \texttt{LPROS()} function returns a new \texttt{mldr} object representing the MLD after resampling. As can be seen from the measurements, this version of the dataset has a few more samples and both \textit{MeanIR} and \textit{SCUMBLE} have decreased, which is positive.

Since all functions corresponding to a resampling algorithm return a \texttt{mldr} object as a result, chaining several methods is quite easy. For example, LPROS creates new samples based on minority label combinations rather than individual labels. A dataset with a high degree of coupling between minority and majority labels would hardly benefit from this algorithm. However, label decoupling can be applied before LPROS by using the REMEDIAL algorithm, as shown below.

\begin{lstlisting}[language=R]
> lpemotions <- LPROS(REMEDIAL(emotions), P=25)
> lpemotions$measures[c("num.instances", "meanIR", "scumble")]
$num.instances
[1] 999

$meanIR
[1] 1.18863

$scumble
[1] 0.00224451
\end{lstlisting}

Both the average imbalance ratio and the coupling index provided by the SCUMBLE metric are much better than those achieved by LPROS alone. 

By default, the functions in \texttt{mldr.resampling} do not use parallelization. Because of this, certain algorithms --- those that compute nearest neighbors --- will take a considerable amount of time. A progress bar is used to tell the user how the process is going, as well as an estimate of the time remaining. Once the method is complete, as in the following example, the resulting MLD is available as with any other function.

\begin{lstlisting}[language=R]
> smemotions <- MLSMOTE(emotions, k=5)
|++++++++++++++++++++++++++++++++++++| 100% elapsed=15m 02s
> smemotions$measures[c("num.instances", "meanIR", "scumble")]
$num.instances
[1] 1247

$meanIR
[1] 1.298585

$scumble
[1] 0.004986872
\end{lstlisting}

Distributing the workload across all the processing cores of the user's machine is a one-step operation, calling the \texttt{setParallel(TRUE)} function. This function checks if the required \texttt{parallel} package is installed, and notifies the user if it is not. Once parallel processing is enabled, any method execution will take advantage of this capability. In the example below, the same MLSMOTE algorithms took about a quarter of the time to run.

\begin{lstlisting}[language=R]
> setParallel(TRUE)
# Parallel computing enabled on all 16 available cores. Use function setNumCores if you wish to modify it
> smemotions <- MLSMOTE(emotions, k=5)  # 4m 11s
...
\end{lstlisting}

Although calling the functions associated with individual methods allows the user to test any of them, a planned experiment will usually involve running several of them. This work can be automated using the \texttt{resample()} function, as shown in the example below. In this case, four resampling algorithms are applied to the same MLD. As can be seen from the output of the function, the first step is to compute several structures that act as a nearest neighbor cache. In this way, the overall running time will be shorter than if each algorithm searched for its own set of neighbors.

\begin{lstlisting}[language=R]
> resample(emotions, 
   c("MLROS", "MLRUS", "MLeNN", "MLSOL"), 
   outputDirectory="~/datasets")
# Calculating structures for dataset musicout , if necessary. Once this is done, algorithms will be applied faster
# Calculating VDM table for dataset musicout
# Time taken (in seconds): 0.0046391487121582
# Calculating neighbors structure for dataset musicout . Once this is done, algorithms will be applied faster
# Time taken (in seconds): 485.649948835373
# Running MLROS on musicout with P = 25
# Time taken (in seconds): 0.0200722217559814
# Running MLRUS on musicout with P = 25
# Time taken (in seconds): 0.0121262073516846
# Running MLeNN on musicout with TH = 0.5 and k = 3
# Time taken (in seconds): 0.33689546585083
# Running MLSOL on musicout with P = 25 and k = 3
# Part 1/3: Neighbors were already calculated. That just saved us a lot of time!
# Part 2/3: Calculating auxiliary structures
# Part 3/3: Generating new instances
# Time taken (in seconds): 3.47672557830811
# End of execution. Generated MLDs stored under directory ~/datasets
# algorithm               time
# 1     MLROS 0.0200722217559814
# 2     MLRUS 0.0121262073516846
# 3     MLeNN   485.991483449936
# 4     MLSOL   489.131313562393
\end{lstlisting}

\begin{figure}[h!]
	\includegraphics[width=\linewidth]{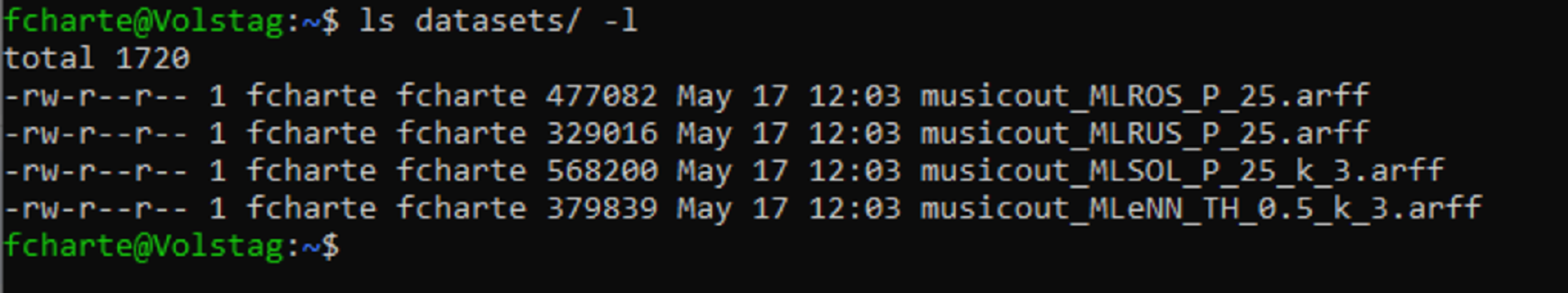}
	\caption{Files produced by the \texttt{resample()} function after running four resampling algorithms over one dataset.}
	\label{Fig.Datasets}
\end{figure}

Since multiple versions of the dataset are created, one for each algorithm, the \texttt{resample()} function will store them in the file system rather than in memory. By default it will use a temporary folder, but the user can specify any path with the \texttt{outputDirectory} parameter. The name of the generated files (see Figure~\ref{Fig.Datasets}) will be a combination of the original MLD, the algorithm and the parameters to run it.

Once all the processed datasets are available, the functionality provided by the \texttt{mldr} package~\cite{Charte:mldr} allows the user to analyze how each algorithm has affected the dataset through a series of metrics and plots.

\section{Conclusions}\label{Sec.Conclusions}

This paper has presented a novel software package, \texttt{mldr.resampling} for the R language, which for the first time provides the user with reference implementations of the most popular multilabel resampling algorithms.

Imbalance is a common problem in multilabel datasets, hence the usefulness of resampling methods as a model-independent approach to overcome this obstacle. However, there are no public implementations of many of the published algorithms. Our \texttt{mldr.resampling} package fills this gap by providing consistent implementations of eleven of these methods. The package is publicly available in CRAN, the official R repository, and users get immediate access to the source code.  In addition, the package is designed for efficiency, minimizing computation by caching nearest neighbor information and parallelizing data processing.

Future plans include the addition of new resampling methods as they become available and with sufficient detail to allow their implementation, extending the functionality already available in the \texttt{mldr} family of packages.

\section*{Acknowledgments}
The research carried out in this study is part of the project ``ToSmartEADs: Towards intelligent, explainable and precise extraction of knowledge in complex problems of Data Science" financed by the Ministry of Science, Innovation and Universities with code PID2019-107793GB-I00 / AEI / 10.13039 / 501100011033.

\bibliographystyle{elsarticle-num}

\end{document}